\documentclass[a4paper]{article}

\usepackage{INTERSPEECH2020}

\usepackage{color}

\title{Attention-based Residual Speech Portrait Model for Speech to Face Generation}
\name{Jianrong Wang$^{1}$, Xiaosheng Hu$^1$,Li Liu$^{2,*}$,Wei Liu$^1$,Mei Yu$^1$,Tianyi Xu$^1$}

\address{
  $^1$College of Intelligence and Computing, Tianjin University, Tianjin 300350, China\\
  $^2$Shenzhen Research Institute of Big Data, the Chinese University of Hong Kong, Shenzhen, Guangdong}
\email{wjr@tju.edu.cn, xiaosheng\_hu@163.com, liuli@cuhk.edu.cn, 602140606@qq.com, yueyouyueyou@gmail.com, tianyi.xu@tju.edu.cn}
\begin{document}
\maketitle
\begin{abstract}
  Given a speaker’s speech, it is interesting to see if it is possible to generate this speaker’s face. One main challenge in this task is to alleviate the natural mismatch between face and speech. To this end, in this paper, we propose a novel Attention-based Residual Speech Portrait Model (AR-SPM) by introducing the ideal of the residual into a hybrid encoder-decoder architecture, where face prior features are merged with the output of speech encoder to form the final face feature. In particular, we innovatively establish a tri-item loss function, which is a weighted linear combination of the L2-norm, L1-norm and negative cosine loss, to train our model by comparing the final face feature and true face feature. Evaluation on AVSpeech dataset shows that our proposed model accelerates the convergence of training, outperforms the state-of-the-art in terms of quality of the generated face, and achieves superior recognition accuracy of gender and age compared with the ground truth.
\end{abstract}

\section{Introduction}

Recently, Audio-Visual Cross-Modal Learning becomes more and more popular \cite{PuttingaFacetotheVoice}, and one of the most interesting topics is to generate face appearances of speakers according to their audio speech. In fact, the research on ``speech to portrait'' has great impacts on our daily life, especially in community security. For example, in the real scene of identifying suspects \cite{Parkhi2015Deep}, when there are no images or appearance features but only audio samples of the suspects, police can use the technology of ``speech to portrait'' to generate the face images, which are similar to suspects' real faces.

In order to explore the feasibility of generating faces from speech, some previous studies have shown that a person’s voice is strongly related to his or her facial structures \cite{Thinkingthevoice,PuttingtheFacetotheVoice}. For example, it was shown in ~\cite{mermelstein1967determination,teager1990evidence} that facial bone, joint structures
and the tissues covering them are closely related to the shape and size of the
organs that produce sound. Meanwhile, genetic factors, biological factors and environmental factors, especially gender \cite{Kotti2008Gender,inproceedings}, age \cite{PtacekAge,Ageestimation,Zazo2018Age} and ethnicity, can largely influence one's voice and face. %Therefore, we can construct a voice portrait model based on Machine Learning or Deep Learning.

Recently, some research works on speech to face generation using deep learning methods have emerged  \cite{ProfilingHumansfromtheirVoice,oh2019speech2face}, among which Speech2Face \cite{oh2019speech2face} is the state-of-the-art (SOTA). It directly generates a speech by Convolutional
Neural Network (CNN) \cite{Krizhevsky2012ImageNet} that learns facial features \cite{merler2019diversity}, and then uses this feature to generate face \cite{ColeSynthesizing}. This method obtains good results, but the CNN structure of speech encoder is complicated, and conventional CNNs may suffer from the gradient vanishing problem. More importantly, due to the natural mismatch between speech and face \cite{singh2019reconstruction}, there is still room to reduce the mismatch and improve the performance.

%%Although there are many interactive factors between voice and face, the mismatch between these two modalities cannot be eliminated completely.
%Due to this intrinsic incompleteness, it is hard to extract face information completely from voice.
%Indeed, this is an unsolved challenge in speech to portrait generation task. However, the existing Speech2Face \cite{oh2019speech2face} model is directly from a speech by CNN network learning facial features, and then use this feature to generate face, does not solve the mismatch between speech and face.

To overcome the shortcomings of previous works and further alleviate the mismatch between speech and face, in this work, we propose a novel Attention-based Residual Speech Portrait Model (AR-SPM) within an encoder-decoder architecture by introducing prior face features, which guides the speech feature to approach to the original face feature as much as possible. Besides, Convolutional Block Attention Module (CBAM) \cite{woo2018cbam} is incorporated into the encoder to extract the key information of the speech and makes the decoder only focus on the face feature, ignoring the background noise. Particularly, we propose a tri-item loss function for encoder, which contains the linear combination of the L2-norm, L1-norm and the negative cosine loss to take into account both errors in values and directions. The decoder loss function is constructed by replacing the cosine loss with the Structural Similarity Index (MS-SSIM) \cite{ZhaoLoss}. The qualitative and quantitative evaluation on AVSpeech dataset \cite{ephrat2018looking} show the superior performance of our proposed method, which outperforms the SOTA results \cite{oh2019speech2face}. An overview of our proposed AR-SPM is shown in Figure \ref{fig:pipeline}.

Overall, our contributions can be summarized as follows:

\begin{itemize}
 \item We propose a novel AR-SPM based on an encoder-decoder architecture. A residual strategy is introduced by incorporating a prior face feature to make the network capture the most representative face features and improve the learning effect of speech portrait model.
 \item We improve the network structure by adding a spatial-channel attention mechanism and constructing a symmetrical face decoder network. Moreover, we innovatively establish a tri-item loss function, which contains L2-norm, L1-norm and negative cosine loss to accelerate the convergence of training.
 \item Experimental results on AVSpeech dataset \cite{ephrat2018looking} show that our model achieves the SOTA results and significantly reduces the cos similarity degree from $40.66^{\circ}$ to $15.20^{\circ}$. 
\end{itemize}

\section{Related Work}
In the literature, three classical methods are used to construct a mapping between speech and face. 
The first method used some attributes classifiers to predict some fixed attributes from speaker’s voice such as gender, age \cite{Zazo2018Age} and nationality. Then it searched the most matching face images automatically based on the prediction results from the local face images database. Lastly, it merged the selected face images using the method in \cite{zhmoginov2016inverting}. However, this speech portrait method has some limits in the accuracy of the attributes classifiers since it needs to be trained in a supervised manner and thus limits the correlation between face and speech within a group of attributes.

The second method that realized a speech portrait model is based on a direct regression using a Deep Neural Network (DNN) \cite{Yu2015Deep}. Specifically, the DNN structure is used to construct the mapping between speech and fixed-resolution RGB face image \cite{duarte2019wav2pix,Goodfellow2014Generative,ReedGenerative,XiaAuxiliary}. However, this method is hard to realize because such a model needs speaker’s original face image as a training label. As a result, the model is easily susceptible to uncontrollable factors in the face image, such as facial expression, head posture, object occlusion, lighting conditions and background, which will greatly affect the regression effect of the model. In addition, the model must learn autonomously and parse out many complex nonlinear transformations \cite{oh2019speech2face}.

The third method is an improvement of the second method by splitting the networks into two parts. The encoder independently learns to extract face-related information from the speech, and the decoder independently learns to restore a standard resolution face image from face feature as in \cite{oh2019speech2face}. Such a model is trained in a self-supervised way, by utilizing the pair of face and speech extracted from videos. The description formula of the network is :
\begin{align}\label{eq:m3}
f = FD(SE(s)),
\end{align}
where $s$ is the speech and $f$ is face image. SE means speech encoder network and FD means face decoder network.
Even this method \cite{oh2019speech2face} is creative and remarkable, it still needs further improvement. Due to the natural mismatch between speech and face \cite{singh2019reconstruction}, there is still room to improve the performance and reduce the mismatch. 

To this end, in this work, we propose a novel AR-SPM to further optimize the SE structure and propose to incorporate a prior face feature to complement the speech feature.

\section{Method}
\begin{figure*}[htbp]
 \centering
 \includegraphics[width=1.0\textwidth]{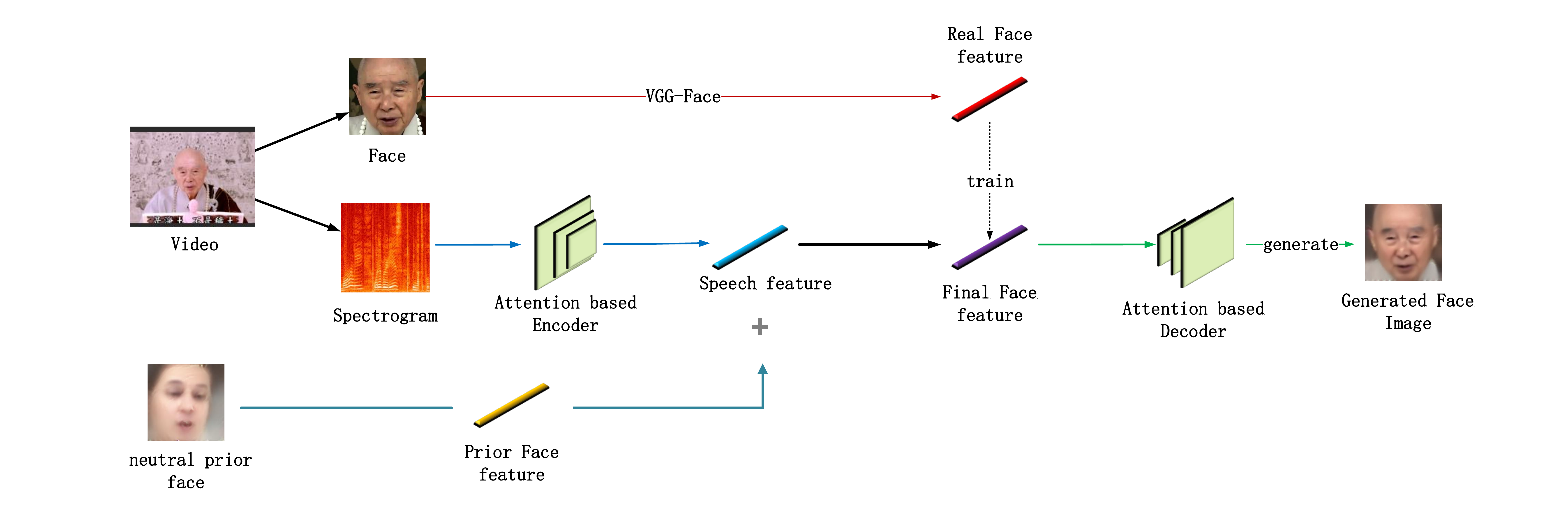}
 \caption{The pipeline of our AR-SPM with the neutral prior face.}
 \label{fig:pipeline}
\end{figure*}

In the following, we describe the details of our proposed AR-SPM (see Figure~\ref{fig:pipeline}), which includes the prior
face features, the CBAM and two new loss functions as well as the structure for SE and FD network, respectively.

%\begin{itemize}
%\item Proceedings will be printed in DIN A4 format. Authors must submit their papers in DIN A4 format.
%\item Two columns are used except for the title section and for large figures that may need a full page width.
%\item Left and right margin are 20 mm each. 
%\item Column width is 80 mm. 
%\item Spacing between columns is 10 mm.
%\item Top margin is 25 mm (except for the first page which is 30 mm to the title top).
%\item Bottom margin is 35 mm.
%\item Text height (without headers and footers) is maximum 235 mm.
%\item Headers and footers must be left empty.
%\item Check indentations and spacings by comparing to this example file (in PDF).
%\end{itemize}
\subsection{Prior face feature}
By introducing the prior face feature, we exploit the idea of the residual to remove the main similar part of the face (i.e., prior face feature), thereby highlighting the small changes depicted by the speech feature. Our AR-SPM converts the speech to face images by a network $\phi$, which is defined as follows.
\begin{align}\label{eq:m3}
\begin{split}
\phi(s)
&=FD\left(SE(s)+f_{prior}\right),
\end{split}
\end{align}
where $\phi(s)$ is the generated face, $s$ means the spectrogram of input speech, and $f_{prior}$ is the prior face feature, which is
calculated before the training stage. Finally, the merged face
feature is upsampled into RGB face image by FD. Two ways are investigated to obtain the final face feature. The first is the
sum of prior face feature and speech feature output from SE,
and the second one is first feeding the sum of the output feature
of SE and prior face feature to a fully connected ($fc$) layer,
and then taking its output as the final face feature.
Recalled that the goal of SE is to learn speech features that mimic facial
features. Motivated by \cite{he2016deep}, SE is
designed to learn the residual, which is the difference between
final face feature and the face prior, instead of learning speech
face feature directly.

A well-defined prior face feature can be approximate to speaker's face feature in many dimensions (e.g., eye contour and lip contour), and contains the main similar part of face, we thus take advantage of adding the prior facial features in the SE to reduce the training difficulties and learn more representative facial features.

\begin{figure}[htbp]
\centering
\setlength{\abovecaptionskip}{-0.1cm}
\includegraphics[width=0.5\textwidth]{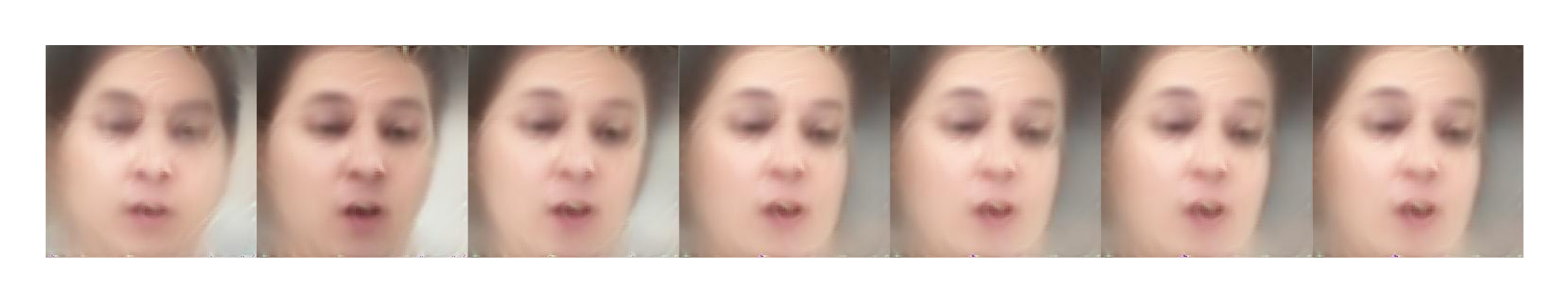}
\caption{The face images restored from neutral prior face-feature by Face Decoder Network when $n$=10, 50, 100, 500, 1000, 5000, 10000.}
\label{fig:prior_neutral}
\end{figure}

Two face priors are investigated in this work. The first one
is neutral face prior, which is the arithmetic mean of
a large gender-balance face image dataset:
\begin{align}\label{eq:neutral}
f_{prior}=\frac{1}{n}\sum_{i=1}^{n} {\rm{VGGFace}}(f)
\end{align}
where $f$ denotes the face image and $n$ is the number of face images. VGGFace denotes the CNN structure in~\cite{Parkhi2015Deep}, which is exploited to extract face features from face images by taking $n$ = 10, 50, 100, 500, 1000, 5000 and 10000. We experimentally find that when $n$ gradually increases, the neutral prior feature tends to converge, indicating that the calculated prior face feature is more representative, as shown in Figure~\ref{fig:prior_neutral} and Table~\ref{tab:prior_gender}. In this work, we finally take $n$ equals 10000.

\begin{table*}[h]
	\centering
	\begin{tabular}{ccccc}
		%\toprule
		\hline
		$n_{1}$& $n_{2}$            &Neutral  $L_{1}(n_{1}, n_{2})$&male $L_{1}(n_{1}, n_{2})$ &female $L_{1}(n_{1}, n_{2})$\\
		%\midrule
		\hline
		$50$& $10$      & 27.861& 43.908 & 36.924\\
		$100$& $50$    & 12.155& 24.302 & 16.318\\
		$500$& $100$    & 10.179& 14.937 & 15.256\\
		$1000$& $500$   & 3.815& 5.059 & 5.797\\
		$5000$& $1000$  & 3.176& 4.254 & 4.505\\
		$10000$& $5000$ & 1.179& 1.967 & 1.913\\
		%\bottomrule
		\hline
	\end{tabular}
	\caption{$L_{1}$ distance between VGGFace face features with different sample face image number $n$. In particular, we calculated them in cases of neutral, male, and female.}
	\label{tab:prior_gender}
\end{table*}

%\subsection{Text font}

%Times or Times Roman font is used for the main text. Font size in the main text must be 9 points, and in the References section 8 points. Other font types may be used if needed for special purposes. It is VERY IMPORTANT that while making the final PDF file, you embed all used fonts! To embed the fonts, you may use the following instructions:
%\begin{enumerate}
%\item For Windows users, the bullzip printer can convert any PDF to have embedded and subsetted fonts.
%\item For Linux/Mac users, you may use \\
   %pdftops file.pdf\\
   %pstopdf -dPDFSETTINGS=/prepress file.pdf
%\end{enumerate}

%\LaTeX users: users should use Adobe Type 1 fonts such as Times or Times Roman. These are used automatically by the INTERSPEECH2020.sty style file. Authors must not use Type 3 (bitmap) fonts.

The second prior face feature is a gender-dependent prior by
assigning two prior face features to male and female, respectively.
To achieve this, a robust classifier network is first needed to predict
the gender based on the audio speech. We establish a lightweight
CNN network, which contains five convolution layers, three
max pooling layers and two fully connected layers to predict the
gender of speakers. It is trained and tested on the VGGFace
dataset, and the gender accuracy reache 97.01\% in test dataset.

Based on the classification results, we calculate the male and female
prior face features using the same method as the neutral prior face feature. The
gender prior features tend to converge as well when the number
of face images increases (see Table 1).

It should be noted that these two prior face features are unbiased on the age due to the AVSpeech dataset contains candidates aged from 20 to 50 years old. Besides, in this work, we do not take into account some potential attributes such as skin color and face shape, which do not affect the speaker's portrait.

\subsection{Convolutional Block Attention Module}
To make our proposed AR-SPM owes an ability to focus on important
features and ignore the irrelevant information, we utilize
a lightweight general-purpose module CBAM~\cite{woo2018cbam}, which can
be easily integrated into an end-to-end CNN architecture.

CBAM contains a channel and a spatial attention, which are embed into both
encoder and decoder. Concerning the channel attention
module, each channel of a feature represents a specialized detector,
so that it can focus on what are important features, while the spatial attention module is used to determine where
are useful features. The channel attention and spatial attention can
be combined together in a parallel or tandem approach. We experimentally
find that ``first channel attention and then spatial
attention” in a tandem manner achieves the best result. Besides, we also tried channel-focused attention mechanism~\cite{HuSqueeze}, and it turns out that CBAM performs better.

As shown in Figure \ref{fig:pipeline}, we apply CBAM in both the SE and FD, so that SE obtain a better feature representation, and FD focus on the feature information of the face, instead of the background noise.  

\begin{table}
	\setlength{\tabcolsep}{0.8mm}{
	\centering
	\begin{tabular}{cccccc}
		%\toprule
		\hline
		Layer & Kernel & Stride & Padding  & Out Padding & Out Channels\\
		%\midrule
		\hline
		Fc1 & 1 & - & - & - & 1000\\
		Fc2 & 1 & - & - & - & 25088\\
		ConvTrans1 & 5 & 2 & 2 & 1 & 512\\
		ConvTrans2 & 5 & 1 & 2 & 0 & 512\\
		ConvTrans3 & 5 & 1 & 2 & 0 & 512\\
		ConvTrans4 & 5 & 1 & 2 & 0 & 512\\
		ConvTrans5 & 5 & 1 & 2 & 0 & 512\\
		ConvTrans6 & 5 & 2 & 2 & 1 & 256\\
		ConvTrans7 & 5 & 1 & 2 & 0 & 256\\
		ConvTrans8 & 5 & 1 & 2 & 0 & 256\\
		ConvTrans9 & 5 & 2 & 2 & 1 & 64\\
		ConvTrans10 & 5 & 1 & 2 & 0 & 64\\
		ConvTrans11 & 5 & 2 & 2 & 1 & 32\\
		Conv1 & 1 & 1 & 0 & - & 3\\
		%\bottomrule
		\hline
	\end{tabular}}
	\caption{The structure of FD network.}
	\label{tab:FD}
\end{table}

\subsection{Encoder-decoder structure}
In this work, we redesigned the SE network as shown in Table ~\ref{tab:SE}, where
the Fc2 layer is a fusion layer for speech feature and prior face feature. The CBAM~\cite{woo2018cbam} is added to the CNN structure. We have also tried the general deep learning structures like Resnets \cite{he2016deep} or Densenets \cite{huang2017densely} as the SE network, but through experiments, we found that these kind of networks are not effective enough as the role of extract face information in speech portrait model. 
We hypothesize that Resnets and Densnets are skip-connected networks, which may introduce noise or silencing fragments, causing longer training steps to extract effective speech feature information.

%We hypothesize that the Densenets can't extract enough voice feature information. Because when extracting the features of the  spectrogram, there may be many silent fragments or noises , so the features extracted from the shallow layer of the network may contain the feature values of the noises or silent fragments, which have not been completely filtered out. Since resnet and Densnet are skip connected networks, this causes the value of noise or silencing fragments to be introduced in addition to the features of the deep layers. So the network needs longer training to extract effective sound feature information.

As for the FD network, we exploit the structure of transposed convolutions in \cite{cole2017synthesizing} and design a network that is symmetrical to the VGGFace model. The structure of FD network is shown in Table~\ref{tab:FD}. To reduce the number of hyperparameters, instead of using a $fc$ layers ($4096 \times 25088$), we first use two $fc$ layers ($4096 \times 1000$ and $1000 \times 25088$) to resize the input feature as a $7 \times 7 \times 512$ feature map that is the same as the last feature map of VGGFace model. Then we use our designed transposed convolutions to upsample the feature map as a $224 \times 224 \times 64$ feature map. Lastly, a $1 \times 1$ convolution is used to convert the feature map to a size $224 \times 224 \times 3$.

\begin{figure*}[htbp]
	\centering
	\setlength{\abovecaptionskip}{-0.1cm}
	\includegraphics[width=0.9\textwidth]{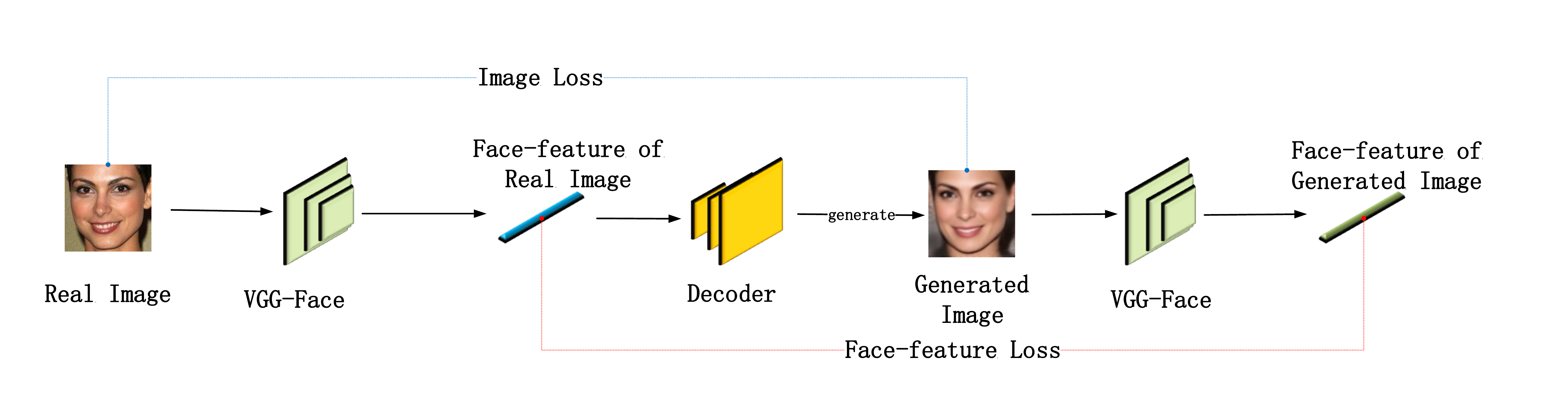}
	\caption{Overview of FD networks.}
	\label{fig:decoder_loss}
	% \vspace{\baselineskip} %表示图与正文空一行
\end{figure*}
\subsection{Proposed loss functions}
In the following, we introduce our proposed loss functions of FD and SE, respectively.

\subsubsection{Face Decoder loss function}
Concerning the loss function of FD networks, we modify the join loss function, which is similar to the loss function in \cite{cole2017synthesizing}. As shown in Figure~\ref{fig:decoder_loss}, our FD loss function is composed of an image loss, which is the error between generated image and original image, and a face feature loss, which is the error between face features of generated image and original image. Such a joint loss function can not only penalize the pixel error between image directly, but also the different of abstract identity. Specifically, we adopt the mixed loss of MS-SSIM and $\ell_{1}$ loss \cite{zhao2015loss} as the image loss:
\begin{align}\label{eq:ms-ssim-l1}
%\begin{split}
L_{image}=\alpha \cdot L_{\rm{MS-SSIM}}+(1-\alpha) \cdot G \cdot L_{\ell_{1}},
%\end{split}
\end{align}
where $\alpha=0.84$ and $G$ is the parameters of Gaussian distribution in MS-SSIM. It considers the influence of resolution, so that it can retain high-frequency information (e.g., image edges or details) but tend to cause brightness changes and color deviations. In order to tackle this problem, we exploit $\ell_{1}$ function to keep brightness and color unchanged. Combining them together can produce a good penalization for image generation task.

In order to pay more attention to face contour in the image rather than the value of the pixel itself, we introduce the cosine similarity loss, which is generally used to measure the difference of two embeddings in the feature space:
\begin{align}\label{eq:cos}
\begin{split}
L_{CS}(A, B) &= 1 - \frac{A \cdot B}{\|A\|\|B\|}
\\&=1 - \frac{\sum_{i=1}^{n} A_{i} \times B_{i}}{\sqrt{\sum_{i=1}^{n}A_{i}^{2}} \times \sqrt{\sum_{i=1}^{n} B_{i}^{2}}}.
\end{split}
\end{align}

Finally, the total loss function of the FD is:
\begin{align}\label{eq:cos}
\begin{split}
L_{Total} &= L_{image}+L_{CS}.
\end{split}
\end{align}

\subsubsection{Speech Encoder loss function}
A tri-item loss function is proposed for SE network. The first item is $\ell_{2}$ loss between the unitized $S_{f}$ (i.e., the output speech feature of SE network) and unitized $F_{f}$ (i.e., the real speaker’s face feature obtained by VGGFace). The second item is $\ell_{1}$ loss between the output of the FD network's first $fc$ layer ($D_{Fc1}$: $\mathbb{R}^{4096} \rightarrow \mathbb{R}^{1000}$), which penalizes the difference of hidden layer features. The third item is Cosine Similarity Loss between the output of the VGGFace third $fc$ layer ($L_{CS}$: $\mathbb{R}^{4096} \rightarrow \mathbb{R}^{2622}$), which penalizes the difference of identity feature. We have also tried the knowledge distillation loss in \cite{oh2019speech2face}, and it shows inferior performance to the Cosine Similarity Loss. Finally, the tri-item loss function is:
\begin{align}\label{eq:tri_item}
\begin{split}
L_{total}&=\lambda_{1}\left\|\frac{F_{f}}{\left\|F_{f}\right\|}-\frac{S_{f}}{\left\|S_{f}\right\|}\right\|_{2}^{2}\\
&+ \lambda_{2} \left\|D_{Fc1}\left(F_{f}\right)-D_{Fc1}\left(S_{f}\right)\right\|_{1}\\
&+ \lambda_{3} L_{CS}\left(V_{Fc3}\left(F_{f}\right), V_{Fc3}\left(S_{f}\right)\right),
\end{split}
\end{align}
where $\lambda_{1}$, $\lambda_{2}$, $\lambda_{3}$ ($\lambda_{1}=1$, $\lambda_{2}=0.04$, $\lambda_{3}=1.2$) are balanced coefficients which make the gradient of loss items within a similar scale. 

Since FD takes part in the calculation of loss function in SE, FD will be trained before SE, and the parameters of FD will be frozen during the training of SE.

\section{Experiment}

\subsection{Datasets and Implementation Details}
The AVSpeech dataset \cite{ephrat2018looking} is used to evaluate our model, which is a large-scale speaking video dataset from YouTube. Besides, we use VGGFace dataset to train our gender prediction network to classify the gender, and 0.15 million videos are used as training set and 8 thousand videos as test set. First, we crop the face image to size $224 \times 224$ from AVSpeech training set to train the FD network, whose structure and implementation details are shown in Table \ref{tab:FD}. Then, we clip and rescale the audio from each AVSpeech video to 6s by concatenating if an audio length is shorter than 6s or cutting if an audio length is longer than 6s. The audio waveform is resampled at 16 kHz and only a single channel is used. Spectrograms are computed by the speaker-independent audio-visual model according to \cite{ephrat2018looking}. 

We use the spectrogram and face-image pairs to train the SE network, whose structure and implementation details are shown in Table \ref{tab:SE}. Our model is implemented by PyTorch 1.1.0 and optimized by Adam \cite{kingma2014adam} with $\beta_{1}=0.5$, $\epsilon=10^{-4}$, the learning rate of 0.001 with the exponentially decay rate of 0.9 at every 2 epochs. We finally train our model with 50 epochs, and the batch size is 16.

\subsection{Evaluation Metrics}

The face image generated by speech portrait model may be interfered with variable face pose, expression, background or noncritical details. Therefore, a series of pixel level evaluation index is not applicable. In this work, we adopt a part of evaluation metrics commonly employed in SOTA work \cite{oh2019speech2face}.

More precisely, we not directly compared the similarity of generated face images with original face images, but compare the similarity of face features extracted from them, because face feature has good robustness on expression, head posture light conditions and background \cite{cole2017synthesizing}. Besides, we compared the speech features from SE network and the corresponding face features from the ground truth face image, using the $L_{1}$ loss, $L_{2}$ loss and Cosine Similarity ($Cos$). When $L_{1}$ and $L_{2}$ tends to 0 or $Cos$ tends to 1, the difference between two images tends to be small or two features are similar in the direction. We also measure the similarity between the feature of the face image generated by the AR-SPM and the corresponding face features from ground truth images using the same three metrics denoted as $L_{1}^{\prime}$, $L_{2}^{\prime}$, $Cos^{\prime}$. Furthermore, we compare the face feature between the original face features and the features of the face generated by the original face features. For convenience, we call it Face-to-Face in this work, which can be regarded as a benchmark for
comparison, since the speech portrait model relies on the face feature from original face image for guidance.

\begin{table}
	\centering
	\begin{tabular}{ccccc}
		%\toprule
		\hline
		Layer & Kernel & Stride & Padding & Out Channels\\
		%\midrule
		\hline
		Conv1 & 7 & 2 & 1 &  64\\
		MaxPool1 & 3 & 2 & 0 & -\\
		Conv2 & 5 & 2 & 1 &  128\\
		MaxPool2 & 3 & 2 & 0 &  -\\
		Conv3 & 3 & 1 & 1 &  256\\
		Conv4 & 3 & 1 & 1 &  512\\
		Conv5 & 3 & 1 & 1 &  512\\
		MaxPool3 & $5\times3$ & $3\times2$ & 0 &  -\\
		CBAM & - & - & - &  -\\
		Fc1 & 1 & - & - &  4096\\
		AvgPool1 & $1\times1$ & - & - &  -\\
		Fc2 & 1 & - & - &  4096\\
		%\bottomrule
		\hline
	\end{tabular}
	\caption{The structure of SE network.}
	\label{tab:SE}
\end{table}

\subsection{Quantitative and Qualitative Analysis}
In the following, we show the quantitative and qualitative results of the proposed FD network and the AR-SPM.
\subsubsection{The Performance of Face Decoder Network}
\begin{figure}[htbp]
\centering
\setlength{\abovecaptionskip}{-0.1cm}
\includegraphics[width=0.5\textwidth]{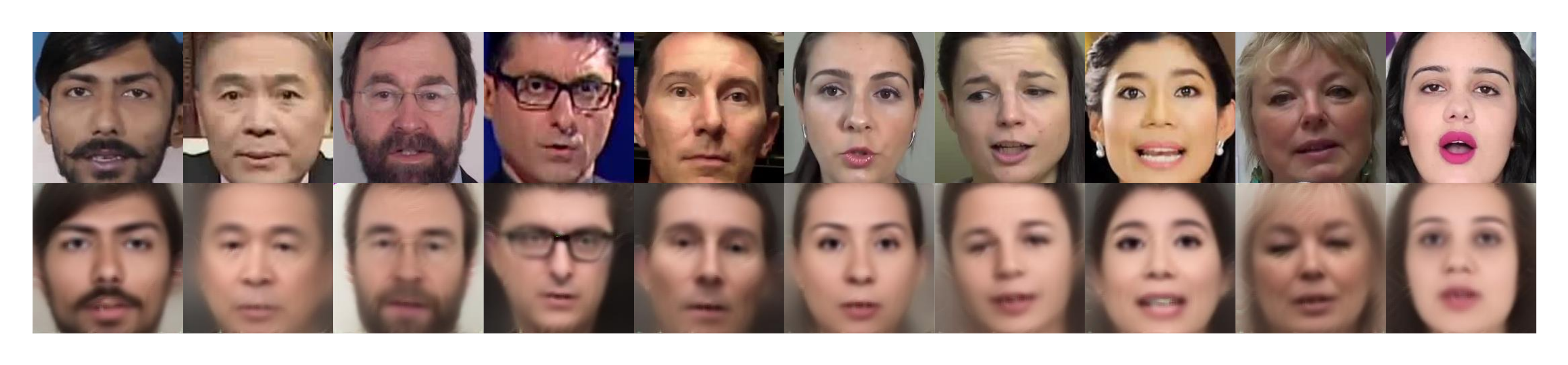}
\caption{The qualitative results of FD network on the AVSpeech dataset. The first row shows the original face image, and the second row shows the image generated by FD.}
\label{fig:f2f}
% \vspace{\baselineskip} %表示图与正文空一行
\end{figure}

We train the FD network on face images from AVSpeech dataset, and show qualitative results of face reconstruction in Figure~\ref{fig:f2f}, we can see that the faces generated by the FD are similar to the ground truth. Besides, since the CBAM attention module is used in our model, the network will concentrate on extracting face features and ignore the background, causing a fuzzy background. 

The quantitative results of FD network are shown in the second row of Table~\ref{tab:feature_similarity}, showing the face feature similarity of face image generated by FD network using the original face image. 

\subsubsection{The Performance of AP-SPM}
To verify the performance of AR-SPM, we train the SE network on spectrogram and face-image pairs from AVSpeech dataset. Let $\rm Model_{neutral}$ and $\rm Model_{gender}$ denote the model with neutral prior face feature and the model with gender prior face feature (given by the proposed automatic gender classifier), trained with data containing both male and female, respectively. $\rm Model_{male}$ and $\rm Model_{female}$ mean the models using the male and female face prior, and training with only male and only female data, respectively. Besides, we denote prior feature models, which use a $fc$ layer to fuse the prior face feature and output speech feature as $\rm Model_{neutral+fc}$ and $\rm Model_{gender+fc}$, respectively.

\begin{figure}[htbp]
	\centering
	\setlength{\abovecaptionskip}{-0.1cm}
	\includegraphics[width=0.4\textwidth]{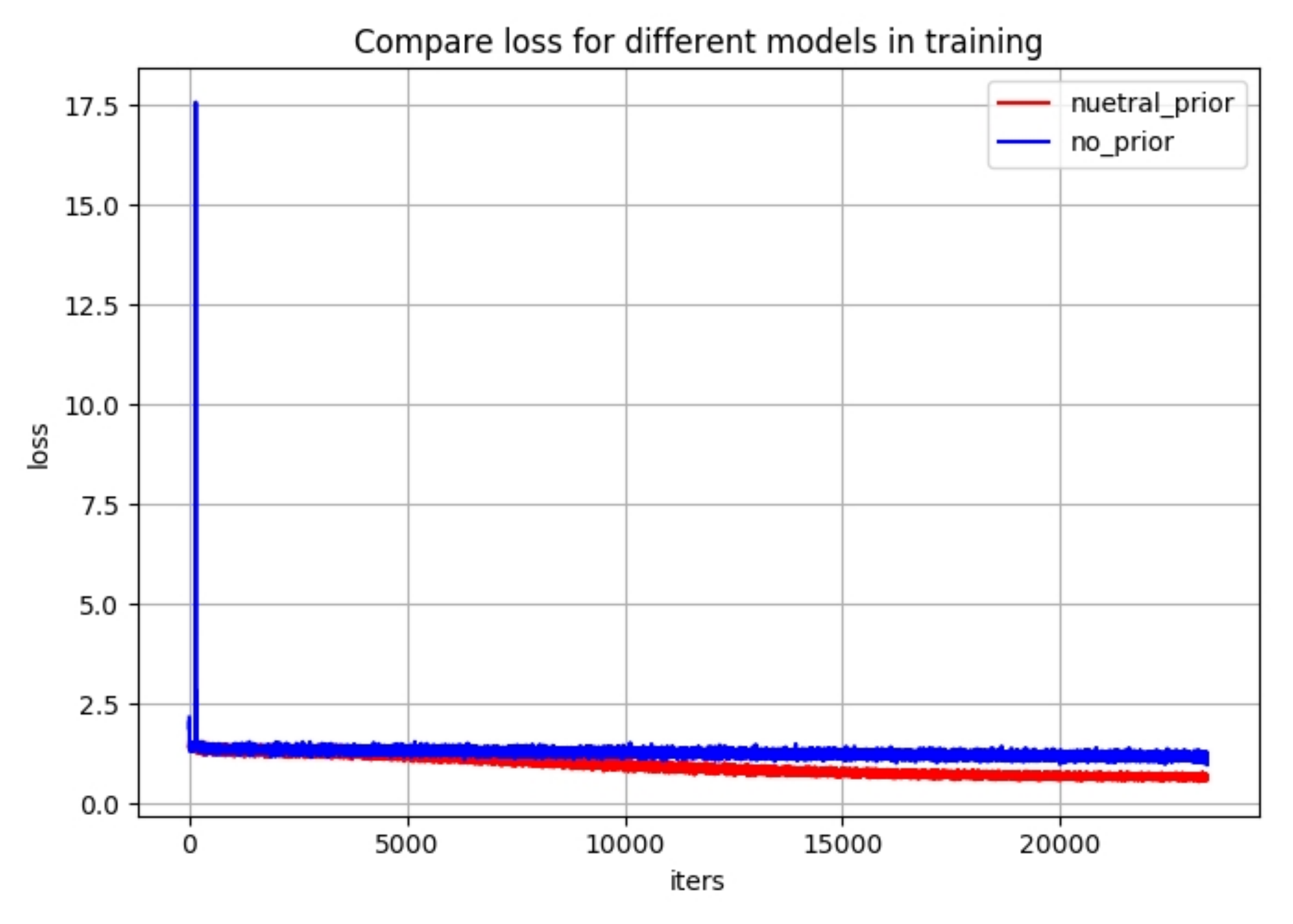}
	\caption{The decline of the loss function, where the blue line denotes the loss of $\rm Model_{non-prior}$, and
		the red line denotes the loss of $\rm Model_{neutral+fc}$.}
	\label{fig:loss}
	% \vspace{\baselineskip} %表示图与正文空一行
\end{figure}

\begin{table*}
	\centering
	\setlength{\tabcolsep}{1mm} 
	\begin{tabular}{ccccccc}
		%\toprule
		\hline
		Model &   $L_{1}$ & $L_{2}$ & $Cos(deg)$ &   $L_{1}^{\prime}$ & $L_{2}^{\prime}$ & $Cos^{\prime}(deg)$  \\
		%\midrule
		\hline
		Face-to-Face      & -        & -      & -   & 97.158  & 1.890   & 14.534 \\
		$\rm Model_{non-prior}$ & 162.957 & 3.176 & 29.191 & 179.594 & 3.499   & 30.231 \\
		$\rm Model_{neutral+fc}$ & \textbf{126.524} & \textbf{2.470}  & \textbf{15.204} & \textbf{144.027} & \textbf{2.809}   & \textbf{25.842} \\
		$\rm Model_{neutral}$    & 143.335 & 2.795 & 27.748 & 148.610 & 2.897   & 25.973 \\
		$\rm Model_{gender+fc}$  & 155.008 & 3.023 & 28.115 & 171.860 & 3.350   & 29.308 \\
		$\rm Model_{gender}$    & 145.192 & 2.831  & 27.994 & 150.900 & 2.941  & 26.104 \\
		$\rm Model_{female}$    & 128.641 & 2.501  & 19.438 & 156.399 &  3.048  & 21.721 \\
		$\rm Model_{male}$    & 130.861 & 2.542  & 19.610 & 152.985 & 2.981  & 21.409 \\
		%\bottomrule
		\hline
	\end{tabular}
	\caption{Quantitative results of the ablation study. The firt column represents the different methods, the second, third and fourth columns represent the L1, L2 and cosine distances between the final face features generated by the SE network and the face features of the original face image, the fifth, sixth and seventh columns represent the L1, L2 and cosine distances between the face features extracted from the face image generated by our model and the face features of the original image.}
	\label{tab:feature_similarity}
\end{table*}

An ablation study is carried out by comparing the similarity between speech feature from SE and the corresponding face features from the original face image by measuring the $L_{1}$, $L_{2}$ distance and $Cos(deg)$ (i.e., cosine similarity in form of degree).
Results are shown in Table~\ref{tab:feature_similarity}, where we notice that $\rm Model_{neutral+fc}$ achieves the best performance among all cases, while $\rm Model_{non-prior}$ has the worst performance. $\rm Model_{gender}$ and $\rm Model_{gender+fc}$ have a slight improvement compared to the result of $\rm Model_{non-prior}$, showing that the use of gender face prior gains less compared with the using of neutral face prior. When we switch to the $\rm Model_{female}$ and $\rm Model_{male}$, we find that they are better than $\rm Model_{gender}$. This is reasonable since they train female and male data separately (i.e., data dependent) and thus make the task easier. More importantly, a fair comparsion in case of the $Cos(deg)$ shows that our method achieves $15.20^{\circ}$, outperforming the SOTA \cite{oh2019speech2face} results $40.66^{\circ}$ by a large margin. Besides,
we see the output feature of $fc$ layer brings evident benefits when
combined with the neutral prior, while it leads to a narrowly worse performance
when combined with the gender prior.

Furthermore, we show the decline of the loss function of $\rm Model_{neutral+fc}$ and $\rm Model_{non-prior}$ by plotting every 1000 steps in batch training in Figure~\ref{fig:loss}. It shows that our $\rm Model_{neutral+fc}$ converges faster than $\rm Model_{non-prior}$, and it is able to avoid the unstable training causing by the abnormal data samples. We can see that two abnormal training loss values are appeared at the beginning of the training stage for $\rm Model_{non-prior}$, while this phenomenon does not appear in the case of $\rm Model_{neutral+fc}$.

\begin{figure}[htbp]
	\centering
	\setlength{\abovecaptionskip}{-0.1cm}
	\includegraphics[width=0.5\textwidth]{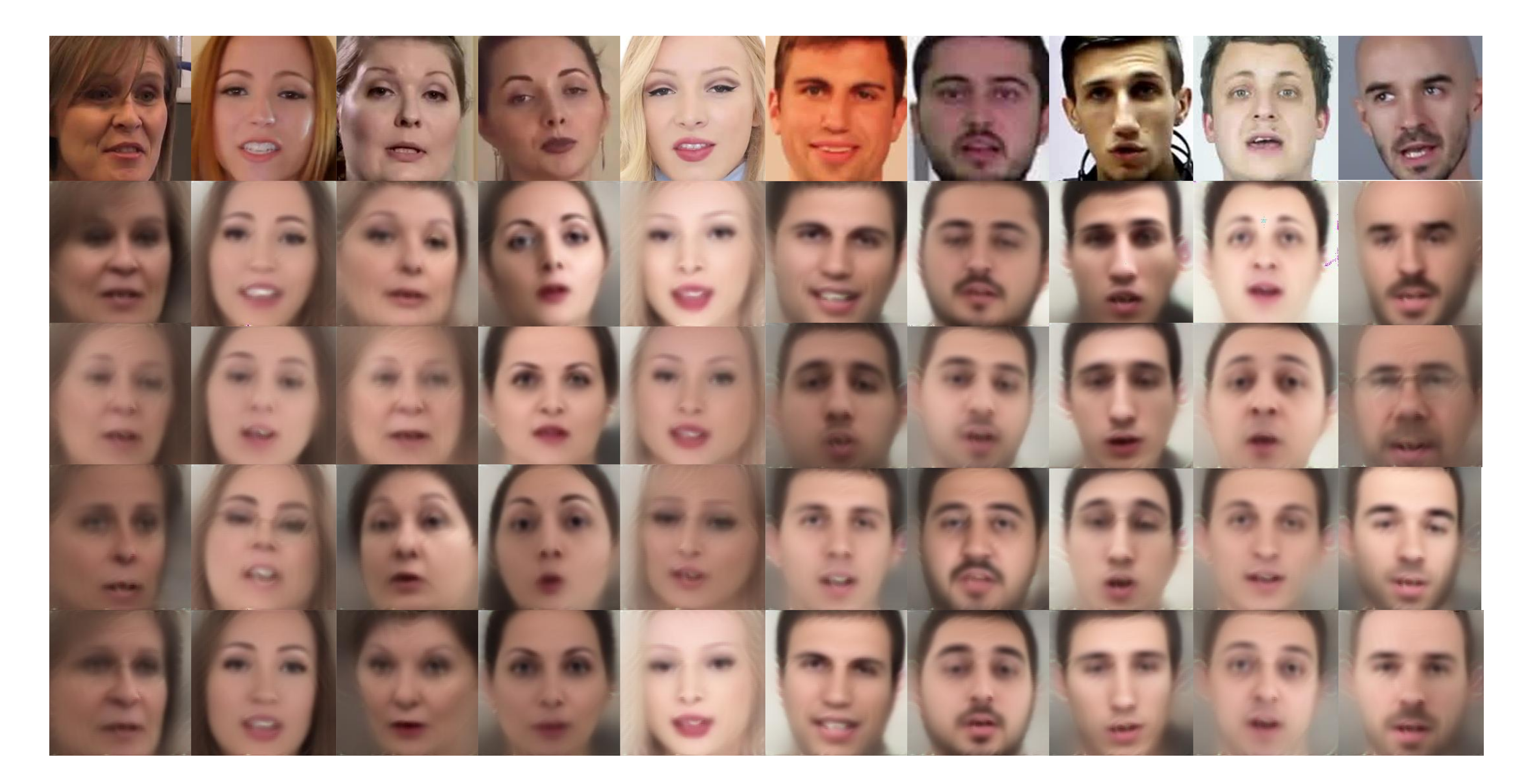}
	\caption{Face generated by the AR-SPM. The 1st row: original face images cropped from video, the 2nd row: the Face-to-Face result, the 3rd row: face images generated from speech by $\rm Model_{non-prior}$, the 4th row: face images generated from speech by $\rm Model_{gender+fc}$, and the 5th row: face images generated from speech by $\rm Model_{neutral+fc}$.}
	\label{fig:s2f}
	% \vspace{\baselineskip} %表示图与正文空一行
\end{figure}

The qualitative results of the ablation study are shown in Figure~\ref{fig:s2f}. We take 10 groups randomly from test set for the demonstration.
% The first row shows the original face image cropped from video, the second row shows the face image reconstruct from the face feature from original face image using Face-to-Face, the third row shows the face image reconstruct from speech using $\rm Model_{non-prior}$ and the fourth row shows the face image reconstruct from speech using $\rm Model_{neutral+fc}$. 
We first notice that face image generated by the Face-to-Face is the most similar to the original face image since it is the reference standard. More importantly, we find that face image generated by the $\rm Model_{neutral+fc}$ is also highly similar to the original face image, showing superior quality to $\rm Model_{non-prior}$. Besides, the $\rm Model_{neutral+fc}$ performs better than $\rm Model_{gender+fc}$, confirming that the neutral prior is more robust than the gender prior in this work. Recalled that by using the CBAM, our model only focus on the face and ignores the background, leading to blur background in generated face images.

\subsubsection{The performance of gender and age recognition}
To further evaluate the quality of our proposed AR-SPM, we conduct a gender and age recognition experiment based on the generated face images by our model. Face++ API \cite{faceplus} is utilized to evaluate the gender and age. Results are shown in Table~\ref{tab:gender_accuracy.} and Table~\ref{tab:age_accuracy}, respectively. Due to $\rm Model_{male}$ and $\rm Model_{female}$ obtains superior performance to $\rm Model_{gender}$ and $\rm Model_{gender+fc}$ as shown in Table \ref{tab:feature_similarity}, we only present the recognition results using  $\rm Model_{male}$ and $\rm Model_{female}$. Naturally, Face-to-Face gives the upper limit of the recognition accuracy. Except that, we find the $\rm Model_{neutral+fc}$ achieves the highest accuracy in all cases. Besides, we can see that  $\rm Model_{male}$ and $\rm Model_{female}$ are narrowly worse than $\rm Model_{neutral+fc}$. It should be mentioned that the AVSpeech dataset is not age-balanced (about 35\% for Young, 55\% for Mid-age and 10\% for Elder). The lower accuracy of the elder age comes from the limited older images in the dataset.

These results show that introducing the idea of the residual by using the neutral prior can improve the efficiency of the model training and further enhance the robustness.  

\begin{table}
\centering
\begin{tabular}{cccc}
%\toprule
\hline
Model & Male & Female & Total\\
%\midrule
\hline
Face-to-Face         & 97.4\%  & 92.7\% & 95.5\%\\
$\rm{Model_{non-prior}}$   & 95.6\%  & 89.8\% & 93.2\%\\
$\rm{Model_{neutral+fc}}$ & \textbf{97.9\%}  & \textbf{89.9\%} & \textbf{94.7\%}\\
$\rm{Model_{female}}$     & 87.7\%  & 93.6\% & 90.1\%\\
$\rm{Model_{male}}$     & 97.4\%  & 68.0\% & 85.8\%\\
%\bottomrule
\hline
\end{tabular}
\caption{The accuracy of gender recognition using the face generated by the proposed AR-SPM.}
\label{tab:gender_accuracy.}
\end{table}

\begin{table}
\centering
\begin{tabular}{ccccc}
%\toprule
\hline
Model & Young & Mid-age & Elder & Total\\
%\midrule
\hline
Face-to-Face         & 76.0\%  & 70.5\% & 46.7\% & 70.3\%\\
$\rm Model_{non-prior}$  & 53.6\%  & 65.5\% & 33.1\% & 57.1\%\\
$\rm Model_{neutral+fc}$ & \textbf{67.6\%}  & \textbf{66.1\%} & \textbf{51.2\%} & \textbf{65.2\%}\\
$\rm Model_{female}$     & 55.5\%  & 70.4\% & 23.5\% & 63.0\%\\
$\rm Model_{male}$     & 64.8\%  & 64.2\% & 40.4\% & 62.9\%\\
%\bottomrule
\hline
\end{tabular}
\caption{The accuracy of age recognition using the face generated by the proposed AR-SPM. Young means under 35 year old, Mid-age means 35 to 65 year old, and Elder means more than 65 years old.}
\label{tab:age_accuracy}
\end{table}

\section{Conclusion}
To alleviate the mismatch between speech and face in the speech-based face generation, we propose a novel AR-SPM based on an end-to-end encoder-decoder structure, which utilizes the additional prior face feature to complement the speech feature in the SE network. Two prior face features (i.e., neutral and gender prior face features) are explored according to the gender. In addition, we re-design the encoder and decoder by incorporating the CBAM into the SE and FD to capture the spatial and channel relationships and suppress noise. Results on AVSpeech dataset show that our proposed AR-SPM accelerates the convergence of training and achieves the SOTA performance. In the future, our model will be explored to eliminate the influence of attributes such as hair and image background on the speaker's face reconstruction, or apply in other application fields, such as preliminary medical image generation or diagnosis from speaker's speech.

\bibliographystyle{IEEEtran}

\bibliography{mybib}

% Generated by IEEEtran.bst, version: 1.13 (2008/09/30)
\begin{thebibliography}{10}
\providecommand{\url}[1]{#1}
\csname url@samestyle\endcsname
\providecommand{\newblock}{\relax}
\providecommand{\bibinfo}[2]{#2}
\providecommand{\BIBentrySTDinterwordspacing}{\spaceskip=0pt\relax}
\providecommand{\BIBentryALTinterwordstretchfactor}{4}
\providecommand{\BIBentryALTinterwordspacing}{\spaceskip=\fontdimen2\font plus
\BIBentryALTinterwordstretchfactor\fontdimen3\font minus
  \fontdimen4\font\relax}
\providecommand{\BIBforeignlanguage}[2]{{%
\expandafter\ifx\csname l@#1\endcsname\relax
\typeout{** WARNING: IEEEtran.bst: No hyphenation pattern has been}%
\typeout{** loaded for the language `#1'. Using the pattern for}%
\typeout{** the default language instead.}%
\else
\language=\csname l@#1\endcsname
\fi
#2}}
\providecommand{\BIBdecl}{\relax}
\BIBdecl

\bibitem{PuttingaFacetotheVoice}
K.~Hoover, S.~Chaudhuri, C.~Pantofaru, M.~Slaney, and I.~Sturdy, ``Putting a
  face to the voice: Fusing audio and visual signals across a video to
  determine speakers,'' 05 2017.

\bibitem{Parkhi2015Deep}
O.~M. Parkhi, A.~Vedaldi, A.~Zisserman \emph{et~al.}, ``Deep face
  recognition,'' in \emph{Proceedings of British Machine Vision Conference
  (BMVC)}, vol.~1, no.~3, 2015, pp. 6--17.

\bibitem{Thinkingthevoice}
P.~Belin, S.~Fecteau, and C.~Bédard, ``Thinking the voice: Neural correlates
  of voice perception,'' \emph{Trends in cognitive sciences}, vol.~8, pp.
  129--35, 04 2004.

\bibitem{PuttingtheFacetotheVoice}
M.~Kamachi, H.~Hill, K.~Lander, and E.~Vatikiotis-Bateson, ``‘putting the
  face to the voice’: Matching identity across modality.'' \emph{Current
  biology : CB}, vol.~13, pp. 1709--14, 10 2003.

\bibitem{mermelstein1967determination}
P.~Mermelstein, ``Determination of the vocal-tract shape from measured formant
  frequencies,'' \emph{The Journal of the Acoustical Society of America},
  vol.~41, no.~5, pp. 1283--1294, 1967.

\bibitem{teager1990evidence}
H.~Teager and S.~Teager, ``Evidence for nonlinear sound production mechanisms
  in the vocal tract,'' \emph{Speech Production and Speech Modelling}, vol.~55,
  pp. 241--261, 1990.

\bibitem{Kotti2008Gender}
M.~Kotti and C.~Kotropoulos, ``Gender classification in two emotional speech
  databases,'' in \emph{International Conference on Pattern Recognition}, 2008.

\bibitem{inproceedings}
M.~Feld, F.~Burkhardt, and C.~Müller, ``Automatic speaker age and gender
  recognition in the car for tailoring dialog and mobile services,'' 01 2010,
  pp. 2834--2837.

\bibitem{PtacekAge}
P.~H. Ptacek and E.~K. Sander, ``Age recognition from voice,'' \emph{Journal of
  Speech \& Hearing Research}, vol.~9, no.~2, p. 273.

\bibitem{Ageestimation}
M.~Bahari, M.~McLaren, H.~Van~hamme, and D.~Van~Leeuwen, ``Age estimation from
  telephone speech using i-vectors,'' \emph{13th Annual Conference of the
  International Speech Communication Association 2012, INTERSPEECH 2012},
  vol.~1, pp. 506--509, 01 2012.

\bibitem{Zazo2018Age}
R.~Zazo, P.~S. Nidadavolu, N.~Chen, J.~Gonzalez-Rodriguez, and N.~Dehak, ``Age
  estimation in short speech utterances based on lstm recurrent neural
  networks,'' \emph{IEEE Access}, vol.~6, pp. 22\,524--22\,530, 2018.

\bibitem{ProfilingHumansfromtheirVoice}
R.~Singh, \emph{Profiling Humans from their Voice}, 01 2019.

\bibitem{oh2019speech2face}
T.-H. Oh, T.~Dekel, C.~Kim, I.~Mosseri, W.~T. Freeman, M.~Rubinstein, and
  W.~Matusik, ``Speech2face: Learning the face behind a voice,'' in
  \emph{Proceedings of the IEEE Conference on Computer Vision and Pattern
  Recognition}, 2019, pp. 7539--7548.

\bibitem{Krizhevsky2012ImageNet}
A.~Krizhevsky, I.~Sutskever, and G.~Hinton, ``Imagenet classification with deep
  convolutional neural networks,'' in \emph{International Conference on Neural
  Information Processing Systems}, 2012.

\bibitem{merler2019diversity}
M.~Merler, N.~Ratha, R.~S. Feris, and J.~R. Smith, ``Diversity in faces,''
  2019.

\bibitem{ColeSynthesizing}
F.~Cole, D.~Belanger, D.~Krishnan, A.~Sarna, I.~Mosseri, and W.~T. Freeman,
  ``Synthesizing normalized faces from facial identity features.''

\bibitem{singh2019reconstruction}
R.~Singh, \emph{Reconstruction of the Human Persona in 3D from Voice, and its
  Reverse}.\hskip 1em plus 0.5em minus 0.4em\relax Berlin: Springer, 2019, pp.
  325--363.

\bibitem{woo2018cbam}
S.~Woo, J.~Park, J.-Y. Lee, and I.~So~Kweon, ``Cbam: Convolutional block
  attention module,'' in \emph{Proceedings of the European Conference on
  Computer Vision (ECCV)}, 2018, pp. 3--19.

\bibitem{ZhaoLoss}
H.~Zhao, O.~Gallo, I.~Frosio, and J.~Kautz, ``Loss functions for image
  restoration with neural networks,'' \emph{IEEE Transactions on Computational
  Imaging}, vol.~3, no.~1, pp. 47--57.

\bibitem{ephrat2018looking}
A.~Ephrat, I.~Mosseri, O.~Lang, T.~Dekel, K.~Wilson, A.~Hassidim, W.~T.
  Freeman, and M.~Rubinstein, ``Looking to listen at the cocktail party: a
  speaker-independent audio-visual model for speech separation,'' \emph{ACM
  Transactions on Graphics (TOG)}, vol.~37, no.~4, pp. 112:1--112:11, 2018.

\bibitem{zhmoginov2016inverting}
A.~Zhmoginov and M.~Sandler, ``Inverting face embeddings with convolutional
  neural networks,'' \emph{arXiv preprint}, p. arXiv:1606.04189, 2016.

\bibitem{Yu2015Deep}
D.~Yu and L.~Deng, ``Deep neural networks,'' 2015.

\bibitem{duarte2019wav2pix}
A.~Duarte, F.~Roldan, M.~Tubau, J.~Escur, S.~Pascual, A.~Salvador, E.~Mohedano,
  K.~McGuinness, J.~Torres, and X.~Giro-i Nieto, ``Wav2pix: speech-conditioned
  face generation using generative adversarial networks,'' in \emph{IEEE
  International Conference on Acoustics, Speech and Signal Processing
  (ICASSP)}, vol.~3.\hskip 1em plus 0.5em minus 0.4em\relax IEEE, 2019, pp.
  8633--8637.

\bibitem{Goodfellow2014Generative}
I.~J. Goodfellow, J.~Pouget-Abadie, M.~Mirza, B.~Xu, D.~Warde-Farley, S.~Ozair,
  A.~Courville, and Y.~Bengio, ``Generative adversarial networks,''
  \emph{Advances in Neural Information Processing Systems}, vol.~3, pp.
  2672--2680, 2014.

\bibitem{ReedGenerative}
S.~Reed, Z.~Akata, X.~Yan, L.~Logeswaran, B.~Schiele, and H.~Lee, ``Generative
  adversarial text to image synthesis.''

\bibitem{XiaAuxiliary}
X.~Xia, R.~Togneri, F.~Sohel, and D.~Huang, ``Auxiliary classifier generative
  adversarial network with soft labels in imbalanced acoustic event
  detection,'' \emph{IEEE Transactions on Multimedia}, pp. 1--1.

\bibitem{he2016deep}
K.~He, X.~Zhang, S.~Ren, and J.~Sun, ``Deep residual learning for image
  recognition,'' in \emph{Proceedings of the IEEE conference on computer vision
  and pattern recognition}, 2016, pp. 770--778.

\bibitem{HuSqueeze}
J.~Hu, L.~Shen, S.~Albanie, G.~Sun, and E.~Wu, ``Squeeze-and-excitation
  networks.''

\bibitem{huang2017densely}
G.~Huang, Z.~Liu, L.~Van Der~Maaten, and K.~Q. Weinberger, ``Densely connected
  convolutional networks,'' in \emph{Proceedings of the IEEE conference on
  computer vision and pattern recognition}, 2017, pp. 4700--4708.

\bibitem{cole2017synthesizing}
F.~Cole, D.~Belanger, D.~Krishnan, A.~Sarna, I.~Mosseri, and W.~T. Freeman,
  ``Synthesizing normalized faces from facial identity features,'' in
  \emph{Proceedings of the IEEE Conference on Computer Vision and Pattern
  Recognition}, 2017, pp. 3703--3712.

\bibitem{zhao2015loss}
H.~Zhao, O.~Gallo, I.~Frosio, and J.~Kautz, ``Loss functions for neural
  networks for image processing,'' \emph{arXiv preprint}, p. arXiv:1511.08861,
  2015.

\bibitem{kingma2014adam}
D.~P. Kingma and J.~Ba, ``Adam: A method for stochastic optimization,''
  \emph{arXiv preprint arXiv:1412.6980}, 2014.

\bibitem{faceplus}
MEGVII, ``Face analyze api,''
  \url{https://www.faceplusplus.com.cn/attributes/}, 2019.

\end{thebibliography}

% \begin{thebibliography}{9}
% \bibitem[1]{Davis80-COP}
%   S.\ B.\ Davis and P.\ Mermelstein,
%   ``Comparison of parametric representation for monosyllabic word recognition in continuously spoken sentences,''
%   \textit{IEEE Transactions on Acoustics, Speech and Signal Processing}, vol.~28, no.~4, pp.~357--366, 1980.
% \bibitem[2]{Rabiner89-ATO}
%   L.\ R.\ Rabiner,
%   ``A tutorial on hidden Markov models and selected applications in speech recognition,''
%   \textit{Proceedings of the IEEE}, vol.~77, no.~2, pp.~257-286, 1989.
% \bibitem[3]{Hastie09-TEO}
%   T.\ Hastie, R.\ Tibshirani, and J.\ Friedman,
%   \textit{The Elements of Statistical Learning -- Data Mining, Inference, and Prediction}.
%   New York: Springer, 2009.
% \bibitem[4]{YourName17-XXX}
%   F.\ Lastname1, F.\ Lastname2, and F.\ Lastname3,
%   ``Title of your INTERSPEECH 2020 publication,''
%   in \textit{Interspeech 2020 -- 20\textsuperscript{th} Annual Conference of the International Speech Communication Association, September 15-19, Graz, Austria, Proceedings, Proceedings}, 2020, pp.~100--104.
% \end{thebibliography}

\end{document}